\documentclass[conference, 10pt]{IEEEtran}

\renewcommand\baselinestretch{0.945}

\usepackage[labelsep=period]{caption}

\usepackage{cite}

\usepackage{soul} 
\usepackage{xcolor}

\usepackage{times}

\usepackage[hidelinks, bookmarks=false]{hyperref}

\usepackage{cmap}
\usepackage{mmap}

\usepackage{lipsum}

\usepackage[]{graphicx}

\usepackage{subfig}

\usepackage{amsmath}
\usepackage{amsfonts}

\usepackage[T1]{fontenc}

\usepackage{balance}

\usepackage{siunitx}

\usepackage{enumitem}

\usepackage{algorithm}
\usepackage{algpseudocode}

\makeatletter
\def\@copyrightspace{\relax}
\makeatother

\begin{document}
	
	\title{High-Performance FPGA Implementation of Equivariant Adaptive Separation via Independence Algorithm for Independent Component Analysis}

	\author{\IEEEauthorblockN{Mahdi Nazemi, Shahin Nazarian, and Massoud Pedram}
	\IEEEauthorblockA{Department of Electrical Engineering, University of Southern California, Los Angeles, CA, USA\\
		\url{{mnazemi, shahin, pedram}@usc.edu}}
	}
		
	\maketitle
	\begin{abstract}
		Independent Component Analysis (ICA) is a dimensionality reduction technique that can boost efficiency of machine learning models that deal with probability density functions, e.g. Bayesian neural networks.
		Algorithms that implement adaptive ICA converge slower than their nonadaptive counterparts, however, they are capable of tracking changes in underlying distributions of input features.
		This intrinsically slow convergence of adaptive methods combined with existing hardware implementations that operate at very low clock frequencies necessitate fundamental improvements in both algorithm and hardware design.
		This paper presents an algorithm that allows efficient hardware implementation of ICA.
		Compared to previous work, our FPGA implementation of adaptive ICA improves clock frequency by at least one order of magnitude and throughput by at least two orders of magnitude.
		Our proposed algorithm is not limited to ICA and can be used in various machine learning problems that use stochastic gradient descent optimization.
	\end{abstract}

	\section{Introduction}
	ICA \cite{ica} is a method for finding components that 
	\begin{itemize}
		\item[--] represent essential structure of data, 
		\item[--] are statistically independent, 
		\item[--] and have non-Gaussian distributions.
	\end{itemize}
	ICA can be used to reduce the dimensionality of input features in order to improve performance of a machine learning model, reduce the memory required for storing features, and enhance visualization.
	Dimensionality reduction is also a beneficial preprocessing step for transforming the original problem into a smaller problem suitable for hardware implementation.
	ICA is used in a wide variety of applications including the ones that deal with electroencephalogram (EEG) and Electrocardiography (ECG) (\cite{ica_app1}, \cite{ica_app2}, \cite{ica_app3}, and \cite{ica_app4}), face recognition \cite{ica_app_face}, predicting stock market prices \cite{ica_app_stock}, mobile phone communications \cite{aapo_book}, etc.

	In problems where underlying distributions of input features remain unchanged, the training phase can be done once in software in order to find model parameters.
	Model parameters are then transferred to the hardware that is required to implement model deployment, a task which is typically much simpler than training.
	However, if underlying distributions of input features change over time, an adaptive method that can track these changes is preferred.
	This requires repeating the training phase in software followed by transferring parameters to hardware after each pass of training, or designing a piece of hardware that is capable of model creation, training, and deployment.
	Because software implementations are usually slower than their hardware counterparts and communication of parameters can be costly, especially for pin-constrained platforms like FPGAs, the main focus of this paper will be on the solution that implements model creation, training, and deployment in hardware.

	In this work, we introduce a modified version of Equivariant Adaptive Separation via Independence (EASI) \cite{easi} algorithm for independent component analysis which is more suitable for hardware implementation.
	The hardware implementation of our proposed algorithm operates at a much higher clock frequency compared to previous implementations, has a considerably higher throughput, and converges faster.

	The remainder of this paper is organized as follows.
	Section~\ref{sec:related} reviews the related work.
	Next, Section~\ref{sec:background} explains problem formulation of ICA followed by EASI algorithm.
	After that, Section~\ref{sec:implementation} and Section~\ref{sec:experiments} discuss the proposed algorithm for ICA, its hardware implementation, and experimental results.
	Finally, Section~\ref{sec:conclusion} concludes the paper.

	\section{Related Work} \label{sec:related}
	
	This section reviews some of the related work on hardware implementation of various ICA algorithms.
	Du et al. \cite{pica} introduce parallel ICA (pICA), an extension to FastICA \cite{fastica}, for hyperspectral image analysis.
	Similar to FastICA, pICA is a nonadaptive method and is therefore incapable of tracking changes in distributions of components.
	Odom \cite{easi_odom} implements EASI on an FPGA using a multi-cycle architecture, 16-bit fixed-point variables, and variable learning rate.
	Mayer-Baese \cite{dsp_fpga} implements EASI on FPGA and shows that it consumes about the same amount of resources compared to generalized Hebbian algorithm PCA, but is more robust and can separate many more signals than the PCA algorithm.
	The hardware implementation of \cite{dsp_fpga} is slow, has a low throughput, and consumes a huge number of FPGA resources.

	\section{Background} \label{sec:background}
	In the standard linear model, input features are modeled as linear combinations of some independent components:
	\begin{equation*}
		{\mathbf{x}}_{m \times 1} = A_{m \times n}{\mathbf{s}}_{n \times 1} \quad m \geq n
	\end{equation*}
	where $\mathbf{x}$ is a column vector of input features, $A$ is the mixing matrix comprised of row vectors ${\mathbf{a}_i}, i = 1, 2, ..., m$, $\mathbf{s}$ is a column vector of random independent components $s_j, j = 1, 2, ..., n$, $m$ is the dimensionality of input features, and $n$ is the dimensionality of independent components.
	%
	%
	Independent components are assumed to be non-stationary, so that different linear models may be in effect at different times.
	%

	The objective of ICA is to find a separation matrix that finds an estimate of independent components, without having any prior information about independent components, $\mathbf{s}$, or the mixing matrix, $A$.
	This can be written as
	\begin{equation*}
		{\mathbf{y}}_{n \times 1} = B_{n \times m}{\mathbf{x}}_{m \times 1}
	\end{equation*}
	where $\mathbf{y}$ is a column vector of estimates of independent components and $B$ is the separation matrix.

	One of the major advantages of ICA over other dimensionality reduction techniques such as PCA and factor analysis is that it deals with non-Gaussian distributions, e.g. heavy-tailed distributions that are common in many real-world datasets.

	Another advantage of ICA is that it finds components that are statistically independent.
	This property has a considerable impact on machine learning models that deal with probability density functions (PDFs).
	For example, in Bayesian neural networks where inputs, weights, and/or outputs are represented by PDFs, a challenging and computationally expensive step is sampling these possibly dependent density functions.
	This problem becomes more complicated when the dependency among distributions involves higher-order statistics (HOS).
	Consequently, if ICA is applied to input features as a preprocessing step, the PDF of each feature in reduced space can be easily sampled independent of other features.

	There are two general ways for estimating independent components.
	The first one is by direct use of HOS and by maximizing a measure of non-Gaussianity.
	The intuition behind these methods is that because sum of two random variables is closer to a Gaussian than original ones, estimated components are independent when a measure of non-Gaussianity is maximized.
	The second one is by indirect use of HOS through nonlinear decorrelation.
	The rationale behind these methods is that if $y_i$ and $y_j$ are independent, any nonlinear transformations $g(y_i)$ and $h(y_j)$ are uncorrelated.
	Therefore, they try to find the separation matrix such that $y_i$ and $y_j$ are uncorrelated and transformed components $g(y_i)$ and $h(y_j)$ are also uncorrelated.
	%
	
	EASI is a gradient-based algorithm that estimates independent components using nonlinear decorrelation.
	EASI has several advantages compared to other algorithms for ICA.
	First, it is an adaptive algorithm which makes it suitable for problems where underlying distributions of input features change over time.
	In problems where adaptivity is not a must, there are superior algorithms such as FastICA, which seeks an orthogonal rotation of whitened data through a fixed-point iteration scheme.
	Second, it is equivariant, i.e. convergence rates, stability conditions, and interference rejection levels depend only on normalized distributions of source signals and are independent of the mixing matrix.
	Third, unlike other methods that require whitening of input features as a preprocessing step, it merges whitening with separation, which improves parallelism.
	And last but not least, the basic operations are computationally efficient since it only requires addition and multiplication.
	%
	
\begin{figure}[t]
		\centering
		\includegraphics[width=0.8\columnwidth]{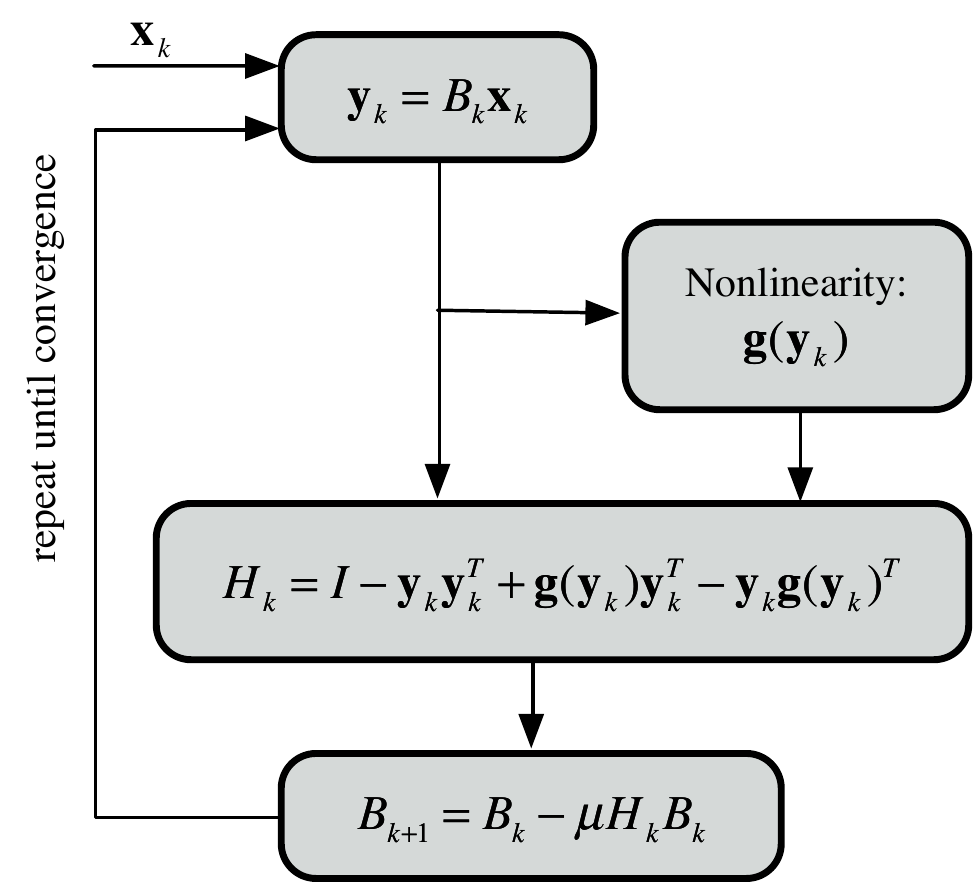}
		\caption{Block diagram of vanilla EASI algorithm.}
		\label{fig:easi_diagram}	
	\end{figure}

	Fig.~\ref{fig:easi_diagram} shows the block diagram of EASI algorithm.
	First, the separation matrix is initialized with random values.
	Then, in each iteration $k$, the separation matrix is multiplied by input features in order to generate output features.
	A nonlinear function $g(.)$ is applied element-wise to output features in order to introduce HOS to the problem.
	The output of nonlinear function and output features are fed to the module that calculates relative gradient $H$ \cite{easi} (aka natural gradient \cite{natural_grad}).
	Finally, relative gradient is multiplied by learning rate $\mu$ to update elements of the separation matrix for the next iteration.
	The same steps are repeated until convergence.

	One of the issues with this algorithm that has caused previous implementations to be relatively slow is the loop-carried dependency due to the separation matrix update.
	Section~\ref{sec:implementation} explains how our design addresses this issue and leads to a hardware that can operate at a higher clock frequency and throughput compared to previous implementations.

	\section{Proposed Algorithm and Hardware Implementation} \label{sec:implementation}
	EASI is a suitable algorithm for adaptive ICA since it implements a stochastic gradient-descent (SGD) optimization.
	Compared to batch gradient-descent, SGD is much faster and more computationally efficient because it updates model parameters based on a single training sample instead of the whole dataset.
	In other words, model parameters for each training sample are strongly affected by the immediately preceding training sample.
	One the other hand, the disadvantage of SGD is that it takes noisier steps towards the minimum which can complicate convergence.

	Mini-batch gradient-descent (MBGD) optimization that takes a few training samples at a time has the best of both worlds in that it is fast and reduces noisy steps \cite{gradient_descent}.
	MBGD applies the same model parameters (separation matrix in this problem) to the training samples within a mini-batch, averages the updated model parameters, and applies them to the next mini-batch.
	It has been shown that MBGD improves performance, because it avoids getting stuck too quickly in local minima, and convergence, because it only relies on a small number of training examples, at the cost of increasing resource consumption proportional to the mini-batch size.
	Our proposed approximation to stochastic gradient descent optimization is similar to MBGD in that it determines model parameters based on a few training samples instead of a single sample, in order to improve performance and convergence.

	Another problem with SGD is that it has trouble making progress where the surface curves more steeply in one dimension than in another \cite{ravine}.
	SGD with Momentum \cite{momentum} is a solution that amplifies gradient for dimensions in which gradients point in the same direction and dampens gradient for dimensions where gradients switch direction in order to improve convergence rate.
	SGD with momentum remembers the update at each iteration, and determines the next update as a convex combination of the gradient and the previous update. 
	Furthermore, it is relatively low-cost to implement SGD with momentum in hardware because it only requires addition and multiplication.

\begin{figure}[t]
		\centering
		\includegraphics[width=0.8\columnwidth]{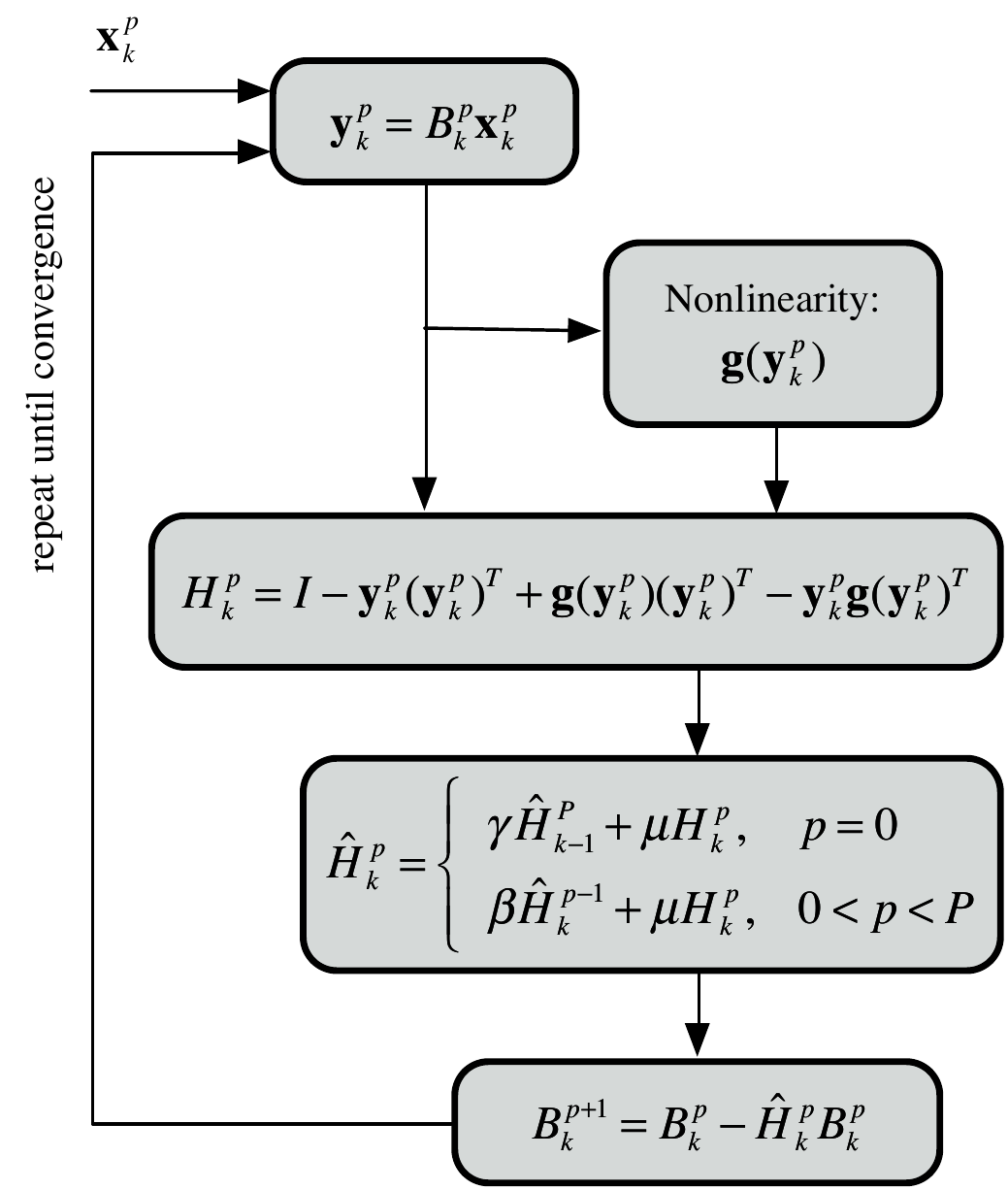}
		\caption{Block diagram of EASI algorithm with SMBGD optimization.}
		\label{fig:smbgd}	
	\end{figure}

	We propose sequential MBGD (SMBGD) for updating the separation matrix.
	SMBGD updates the relative gradient matrix according to Eq.~\ref{eq:smbgd}.
	\begin{equation}
		\hat{H}^{p}_k = 
		\begin{cases}
			\gamma \hat{H}^{P}_{k - 1} + \mu H^{p}_k, & p = 0 \\
			\beta \hat{H}^{p - 1}_k + \mu H^{p}_k,    & 0 < p < P
		\end{cases}
		\label{eq:smbgd}
	\end{equation}
	In this equation, $k$ is the mini-batch index, $p$ is the training sample index within each mini-batch, $P$ is the number of training samples in each mini-batch (mini-batch size), and $\gamma$, $\beta$, and $\mu$ are hyperparameters of the model.
	$\gamma$ is the coefficient that incorporates momentum using gradients in previous mini-batches and $\beta$ is the coefficient that relates various training samples within a mini-batch.
	Note that for the first mini-batch, $\gamma$ is set to zero.
	Fig.~\ref{fig:smbgd} shows the block diagram of EASI with SMBGD.
	To avoid confusion, the following details are removed from the diagram:
	$p$ is incremented in each loop iteration and when $p = P$, $p$ is reset to zero, $\hat{H}^{p}_k$ is reset to a zero matrix, and $k$ is incremented.

	The advantages of SMBGD are threefold.
	First of all, it gets around the loop-carried dependency, which is favorable for a pipelined architecture.
	The pipelined implementation increases clock frequency by breaking down combinational logic into smaller pieces and increases  throughput of the circuit.
	Additionally, in contrast to MBGD that consumes multiple identical hardware resources, SMBGD allows training samples to enter the pipeline one after another, hence significantly decreasing resource consumption compared to MBGD.
	Second, it calculates a weighted average of model parameters which improves convergence similar to MBGD.
	The difference though is that SMBGD multiplies exponentially decaying weights to model parameters found in a mini-batch in order to accentuate more recent samples and account for adaptivity.
	Third, it incorporates a momentum term which further improves convergence rate.
	In problems where underlying distributions change smoothly, larger values of $\gamma$ speed up convergence.
	On the other hand, if distributions change rapidly over time, a lower value of $\gamma$ dampens the effect of previous gradients and puts a higher weight on current samples.

	It should be noted that a pipelined implementation for SGD/MBGD increases resource consumption without considerable improvement in the throughput.
	The reason is that because of the data dependency explained earlier in Section~\ref{sec:background}, the pipeline has to be stalled until the separation matrix is updated and the next training sample is fed.
	On the other hand, SMBGD allows a new training sample to be read at each clock cycle and fed to the pipeline.
	Additionally, in contrast to popular implementations of MBGD, SMBGD does not lead to a linear increase in resource consumption due to its compatibility with pipelined implementations.
	While MGDB is beneficial in platforms such as GPUs, our proposed update rule is preferable in platforms like FPGAs.

	We should mention that SMBGD is not limited to EASI algorithm and in fact, can be used in various machine learning problems that implement some flavor of SGD.

	Section~\ref{sec:experiments} studies the effect of proposed update rule on convergence rate and compares the hardware implementation of enhanced EASI with existing implementations.

	\section{Experimental Results And Discussion} \label{sec:experiments}
	
	\subsection{Performance of Proposed Algorithm}
	In order to compare convergence rate of the proposed update rule with SGD, we run multiple instances of the same separation problem using different random initial values for the separation matrix.
	The number of iterations required for convergence are then averaged across different simulations and compared for the two algorithms.
	Based on our simulations, SGD requires an average of 4166 iterations for convergence while SMBGD requires 3166 iterations.
	As a result, SMBGD improves convergence rate by about 24\%.

	\subsection{FPGA Implementation} \label{subsec:implementation}
	In order to implement our pipelined design including SMBGD on FPGA, we use Chisel \cite{chisel} to create parameterized building blocks required for the design, such as 32-bit floating-point operations, vector-vector outer product, matrix-matrix addition/subtraction/multiplication, and matrix-vector multiplication.
	Chisel is a hardware construction language embedded in Scala \cite{scala} that allows hardware design in a higher level of abstraction by providing object orientation, functional programming, parameterized types, and type inference.

	We use a cubic function in order to implement nonlinearity efficiently.
	While previous implementations use $tanh$ function that is expensive in terms of hardware cost, a cubic function requires fewer additions and multiplications.
	Note that the choice of a nonlinear function, that only requires addition and multiplication, does not affect clock frequency of the pipelined circuit, but only affects the number of logic elements and DSPs consumed on FPGA.
	Simpler nonlinear functions such as rectified linear units (ReLU) that are popular in neural networks may be applied to this problem to further reduce the number of consumed resources on FPGA.

	There is a computation and storage overhead for implementing SMBGD which is quadratic in the number of dimensions in reduced space (because $H$ is a square matrix with dimensionality $n$).
	This overhead is negligible compared to overall complexity of EASI algorithm.
	However, for problems where convergence rate is less important and resources on FPGA are scarce, SMBGD ca be implemented without the momentum term.

	As stated earlier, a fair comparison of our work with previous work is hard because our work uses 32-bit floating point variables and operations and a different nonlinear function.
	In order to compare different implementations fairly, we use the same architecture as the one used in \cite{dsp_fpga}, but use a cubic nonlinear function and 32-bit floating point variables instead.
	Both architectures are synthesized using Quartus Prime Lite Edition on a Cyclone V 5CSEMA5F31C6 FPGA.
	Table~\ref{tbl:results} compares different parameters of EASI with SGD and EASI with SMBGD for a problem where the number of input dimensions is $m = 4$ and the number of output dimensions is $n = 2$.

	\begin{table}[h]
		\caption{EASI with SGD vs. EASI with SMBGD for $m = 4$ and $n = 2$.}
		\label{tbl:results}
		\centering
		\begin{tabular}{l c c}
			Parameters                        & EASI with SGD          & EASI with SMBGD \\
			\hline
			Clock Frequency (\si{\mega\Hz})   & 4.81                   & 55.17 \\
			Throughput (MIPS)                 & 4.81                   & 717.21 \\
			Adaptive Logic Modules            & 12731                  & 10350 \\
			DSPs                              & 42                     & 42 \\
			Registers (bits)                  & 160                    & 3648
		\end{tabular}
	\end{table}
	
	It can be seen that EASI with SMBGD operates at a 11.46 times faster clock frequency, has a 149.11 times higher throughput, but causes a 22.8 times increase in register consumption due to the overhead of pipeline registers.
	It should be noted that while the clock frequency will remain the same for various values of $m$ and $n$, the throughput is proportional to the number of pipeline stages which is equal to $10 + \log_2{mn}$.
	%
	

	\balance

	\section{Conclusion} \label{sec:conclusion}
	In this work, we proposed the stochastic mini-batch gradient descent optimization technique for independent component analysis using the EASI algorithm.
	EASI with SMBGD is suitable for a pipelined implementation and improves both clock frequency and convergence by using mini-batches and introducing a momentum term.
	SMBGD is not limited to EASI and can be used in various machine learning algorithms that use stochastic gradient descent optimization.
	Our implementation of the EASI algorithm improves clock frequency 11.46 times, throughput 149.11 times, and convergence by 24\%.
	While this is a big step towards implementing independent component analysis in hardware, limited resources available on FPGAs restrict scalability of hardware implementation.
	Although this issue can be simply addressed by using multiple FPGAs that work in parallel, alternative solutions that achieve this goal in a fundamentally different manner remain an interesting research direction. 

	\section*{Acknowledgements}	
	The authors would like to thank other members of the VINE (Variational Inference-based Bayesian Neural Network Engine) project team, Shuang Chen and Luhao Wang, for helpful discussions.
	This research was sponsored in part by a contract from DARPA's Microsystems Technology Office.

	\bibliographystyle{IEEEtran}
	\bibliography{IEEEabrv,easi}
	
\end{document}